\documentclass[10pt,twocolumn,letterpaper]{article}

\usepackage{3dv}
\usepackage{times}
\usepackage{epsfig}
\usepackage{graphicx}
\usepackage{amsmath}
\usepackage{amssymb}
\usepackage{float}
\usepackage{subfigure}
\usepackage{threeparttable}
\usepackage{multirow}
\usepackage{array,ragged2e}

\usepackage[pagebackref=true,breaklinks=true,letterpaper=true,colorlinks,bookmarks=false]{hyperref}

\threedvfinalcopy 

\def\threedvPaperID{44} 

\ifthreedvfinal\fi
\begin{document}

\title{UnrealStereo: Controlling Hazardous Factors to Analyze Stereo Vision}

\author{Yi Zhang$^{1}$\thanks{Indicates equal contributions.}, Weichao Qiu$^{1*}$, Qi Chen$^{1}$, Xiaolin Hu$^{2}$, Alan Yuille$^{1}$\\
$^{1}$Johns Hopkins University, $^{2}$Tsinghua University \\
\{\tt\small yzh,wqiu7,qchen42,alan.yuille\}@jhu.edu, {\tt\small xlhu@mail.tsinghua.edu.cn}
}

\maketitle

\begin{abstract}

A reliable stereo algorithm is critical for many robotics applications. But textureless and specular regions can easily cause failure by making feature matching difficult. Understanding whether an algorithm is robust to these hazardous regions is important. Although many stereo benchmarks have been developed to evaluate performance, it is hard to quantify the effect of hazardous regions in real images because the location and severity of these regions are unknown. In this paper, we develop a synthetic image generation tool enabling to control hazardous factors, such as making objects more specular or transparent, to produce hazardous regions at different degrees. The densely controlled sampling strategy in virtual worlds enables to effectively stress test stereo algorithms by varying the types and degrees of the hazard. We generate a large synthetic image dataset with automatically computed hazardous regions and analyze algorithms on these regions. The observations from synthetic images are further validated by annotating hazardous regions in real-world datasets Middlebury and KITTI (which gives a sparse sampling of the hazards).  Our synthetic image generation tool is based on a game engine Unreal Engine 4 and will be open-source along with the virtual scenes in our experiments. Many publicly available realistic game contents can be used by our tool to provide an enormous resource for development and evaluation of algorithms.

\end{abstract}

\section{Introduction}

Stereo algorithms benefit enormously from benchmarks~\cite{scharstein2002taxonomy}. They provide quantitative evaluation to encourage competition and track progress. Despite great progress over the past years, many challenges still remain unsolved, such as transparency, specularity, lack of texture and thin objects. These image regions are called hazardous regions~\cite{zendel2015cv} because they are likely to cause the failure of an algorithm.  These regions are sometimes small, uncommon and do not have a big impact on overall performance, but critical in the real world. For example, a street light is a thin object and covers a small region of an image, but missing it could be a disaster for autonomous driving.  

Images in the real world contain different degrees of hazardous factors, for example, images in KITTI dataset~\cite{menze2015object} contain specular windshields or dark tunnels. In order to better study algorithm robustness, images were captured on extreme weather conditions \cite{meister2012outdoor} or through rendering~\cite{peris2012towards,ros2016synthia}. But these images can only be sparse samples of different hazardous degrees. Even though it is possible to collect a huge dataset with enormous degrees of different hazards, the size of it would be very large making labeling hazardous regions of these images prohibitively expensive. 

\begin{figure}
  \includegraphics[width=\columnwidth]{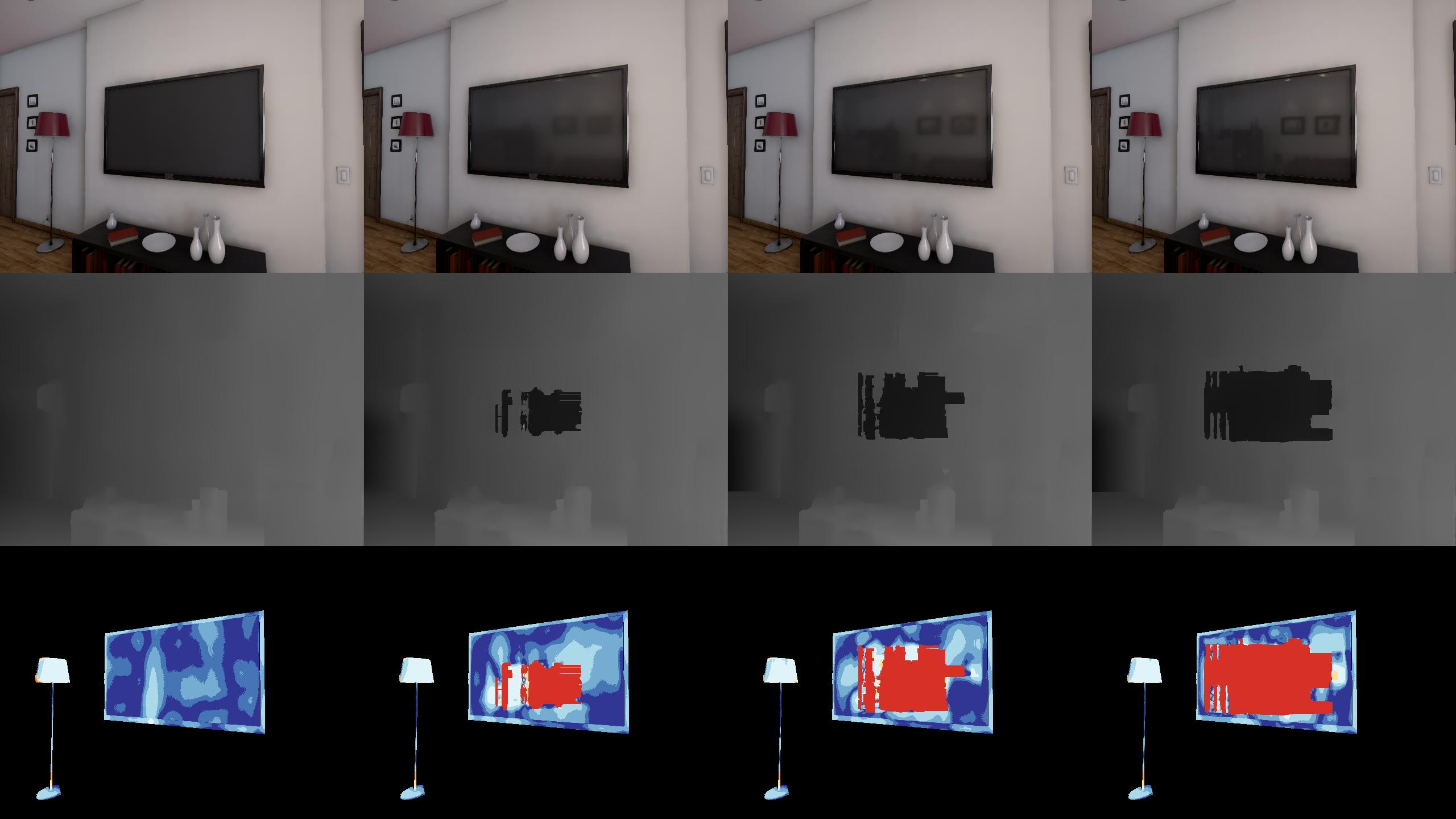}
	\caption{\label{fig:specularity} Different levels of specularity of the TV, from top to bottom are input image, disparity estimation and error compared with ground truth, the error is only computed for the specular regions. The visual difference in the first row is subtle, but is a very big challenge for  state-of-art methods~\cite{chakrabarti2015low}. Best seen in color.}
\end{figure}

To address the problem of thoroughly testing stereo algorithms, we develop a data generation tool for researchers to precisely control {\it hazardous factors}, e.g. material properties, of a virtual scene and produce their own images. For example, in Fig.~\ref{fig:specularity}, we use it to vary the degree of specularity and show how this impacts the performance of a state-of-art stereo algorithm~\cite{chakrabarti2015low}. More generally, our approach enables us to follow the standard strategy in scientific research which changes variables separately and systematically and study their impact. 

In particular, we use this technique in our paper to study the relationship between hazardous factors and algorithm performance to understand the robustness of an algorithm. Adversarial attack~\cite{Xie2017-md,Nguyen2015-ul} is another popular approach to understand model robustness. It requires the model to be differentiable and is mostly applied to deep neural networks. Since the hazardous factors are well understood in binocular stereo~\cite{zendel2015cv}, we are able to study model robustness by controlling the hazardous factors which is more systematical.


In Fig.~\ref{fig:specularity}, the small perturbation of images is done by changing material property, instead of from back-propagation, this perturbation is easy to find and be validated in the real world. The discovery from synthetic images can be validated using real images, and this validation only requires a small amount of test images (hence avoiding the need for excessive annotation of real images). In our diagnosis experiment, after analyzing the impact of individual hazardous factor, we also validate our result on real-world datasets with annotated images. 

In this paper, we use our synthetic image generation tool to study the effect of four important hazardous factors on stereo algorithms. These hazardous factors are chosen to violate some of the basic assumptions of traditional stereo algorithms. For example, specular and transparent surfaces violate the brightness consistency constraint, which assume that the intensity properties of corresponding points are similar (because specularity means that the intensity of a surface point will depend on the viewpoint). Although these hazardous factors are well-known to the community, there have been few attempts at quantitative evaluation of the impact of individual factor due to challenges of annotating these factors. We were inspired by the theoretical framework to analyze hazardous factors proposed by Zendel \etal~\cite{zendel2015cv}, but their framework requires a lot of manual annotation of hazardous regions of images. Our tool can produce these hazardous regions masks automatically, making their theoretical framework practical.


To summarize, we develop a data generation tool called UnrealStereo and use it to stress test stereo algorithms. The main contributions of our paper are as follows: Firstly, we provide a tool enabling researchers to control the hazardous factors in a virtual environment to analyze stereo algorithms. Secondly, hazardous regions are automatically determined in our framework, making the theoretical framework in \cite{zendel2015cv} practical. Third, we control the hazardous factors to show the characteristics of different stereo methods and validate our result on annotations of Middlebury and KITTI dataset. Our tools are open source and will be made available to the community.

\section{Related Work}

\subsection{Robustness Evaluation for Stereo Vision}
Many stereo datasets have been created for training and evaluating stereo algorithms. The Middlebury stereo dataset~\cite{scharstein2002taxonomy,scharstein2003high,hirschmuller2007evaluation,scharstein2014high} is a widely used indoor scene dataset, which provides high-resolution stereo pairs with nearly dense disparity ground truth. The KITTI stereo dataset~\cite{geiger2012we,menze2015object} is a benchmark consisting of urban video sequences where semi-dense disparity ground truth along with semantic labels are available. Tanks and Temples~\cite{Knapitsch:2017:TTB} and ETH3D~\cite{schoeps2017cvpr} are proposed recently as benchmarks for multi-view stereo. Besides these most commonly used ones, \cite{Zendel_2017_CVPR} makes a detailed summary of existing stereo datasets. Due to demand of complex equipment and expensive human labor, real-world datasets usually have relatively small sizes. And the uncertainty in measurements imposes a constraint on the ground truth accuracy of real-world datasets. Furthermore, it is not easy to control hazardous factors in real-world settings. 

Many stereo benchmarks provide scene variation to understand the robustness of stereo algorithms. Middlebury~\cite{hirschmuller2007evaluation,scharstein2014high} provide scenes with varying degrees of illumination and exposure. Neilson \etal~\cite{neilson2008evaluation} provide synthetic data with varying texture, levels of noise and baselines. Tsukuba dataset~\cite{peris2012towards} provides the same synthetic video scene with four different illuminations. In the HCI/Bosch robustness challenge~\cite{meister2012outdoor}, images on challenging weather were captured. In order to test algorithm in different conditions in a controlled way, lab setup based on toys and robotics arm is created~\cite{borji2016ilab} to control hazardous factors, but the images are very different from normal conditions. \cite{morales2009robustness} evaluated the robustness of stereo algorithms against differing noise parameters. Haeusler \etal~\cite{haeusler2013synthesizing} designed cases for typical stereo failure using non-realistic synthetic 2D patterns without an underlying 3D scene. 

Taking the average of pixel errors at full image is not enough for performance evaluation~\cite{kostkova2003dense}. \cite{scharstein2002taxonomy} proposes region specific evaluations for areas of textureless, disparity discontinuities and occlusion. The HCI stereo metrics~\cite{honauer2015hci} focus on disparity discontinuities, planar surfaces, and fine structures. CV-HAZOP~\cite{zendel2015cv} proposes the idea of analyzing hazardous factors in an image. Their method requires manually annotating risk factors, such as specular areas, from images, which is difficult to perform and hard to scale up. Our synthetic pipeline can automatically identify these hazardous regions, enables large-scale analysis. The ability to control the severity of hazardous factors also helps us to better understand the weakness of an algorithm.

\subsection{Synthetic Dataset for Computer Vision}

Synthetic data has attracted a lot of attention recently, because of the convenience of generating large amounts of images with ground truth. And the progress of computer graphics makes synthesizing realistic images much easier. Synthetic data have been used in stereo~\cite{peris2012towards,butler2012naturalistic,haeusler2013synthesizing,ros2016synthia,mayer2016large}, optical flow~\cite{barron1994performance,butler2012naturalistic}, detection~\cite{qiu2016unrealcv,tremblay2018training} and semantic segmentation~\cite{richter2016playing, ros2016synthia, gaidon2016virtual,tsirikoglolu2017procedural}. Images and ground truth are provided in these datasets, but the virtual scenes are not available to render new images or change the properties of these scenes. Instead of constructing proprietary virtual scenes from scratch, we use game projects that are publicly available in the marketplace. Our tool enables tweaking virtual scenes, e.g. by varying the hazardous factors in virtual experiments, to generate more images and ground truth. Many virtual scenes constructed by visual artists in the marketplace can be used. Unlike Sintel~\cite{butler2012naturalistic} and Flyingthing3D~\cite{mayer2016large}, our approach utilizes more realistic 3D models arranged in real-world settings. 





\section{Hazardous Factor Analysis}

Most of stereo algorithms can be formulated in terms of minimizing an objective function w.r.t disparity $d$,
\begin{equation}
E(d) = \sum_{\boldsymbol{p}} E_{d}(d(\boldsymbol{p})) + \lambda \sum_{\boldsymbol{(p,q) \in \mathcal{C}}} E_{s}(d(\boldsymbol{p}), d(\boldsymbol{q}))
\end{equation}
where the data term $E_{d}$ usually represents a matching cost and the smoothness term $E_{s}$ encodes context information within a support region $\mathcal{C}$ of pixel $\mathbf{p}$ ($\mathbf{q}$ is a pixel in $\mathcal{C}$). Local stereo methods~\cite{geiger2010efficient,ma2013constant} do not have a smoothness term and utilize only local matching cues. Global methods~\cite{hirschmuller2005accurate, yamaguchi2014efficient,zbontar2015computing,chakrabarti2015low} incorporate the smoothness priors on neighboring pixels or superpixels in the smoothness term.

The success of these methods relies on some basic assumptions hold for the scene they encounter. First, to do correspondence between binocular image pairs, image patches of the projection of the same surface should be similar which requires Lambertian surface assumption and the single image layer assumption. Second, the local surface should be well-textured for matching algorithms to extract feature. Third, the smoothness term in global method functions under the assumption that the disparity vary slowly and smoothly in space. However, these assumptions can easily be broken in real world scenarios. For example, the first assumption does not hold for specular surface which is not Lambertian and transparent surfaces which would create multiple image layers. Textureless objects are everywhere such as white walls and objects under intense lighting. Besides, smoothness assumption does not hold for regions with many jumps in disparity, e.g. fences and bushes. 

Since the aforementioned factors often break the assumptions of most stereo methods, we call them \textit{hazardous factors} following~\cite{zendel2015cv}. Special efforts have been made to resolve these difficulties in recent years. Yang \etal~\cite{yang2008near} proposed an approach which replaces estimates in textureless regions with planes. Nair \etal~\cite{nair2015reflection} derive a data term that explicitly models reflection. G\"uney \etal~\cite{guney2015displets} leverage semantic informations and 3D CAD models to resolve stereo ambiguities caused by specularity and no texture. An end-to-end trained DCNN based algorithm~\cite{mayer2016large} performs well on specular regions of KITTI stereo 2015 after finetuning on the training set. 

Evaluating stereo algorithms under different hazardous factors on real data is highly inconvenient, because hazardous regions 1) require annotation by human labor and 2) can hardly be controlled. To this end, we develop a synthetic data generation tool for systematic study of hazardous factors.

For the rest of this section, we first describe the data generation tool UnrealStereo. Then we vary the hazardous factors to produce hazardous regions to stress test state of the art stereo algorithms. Finally, hazardous regions are computed for images rendered from realistic 3D scenes to analyze the impact of each hazardous factor.

\subsection{UnrealStereo Data Generation Tool}

\begin{figure*}
	\centering
    \includegraphics[width=0.8\linewidth]{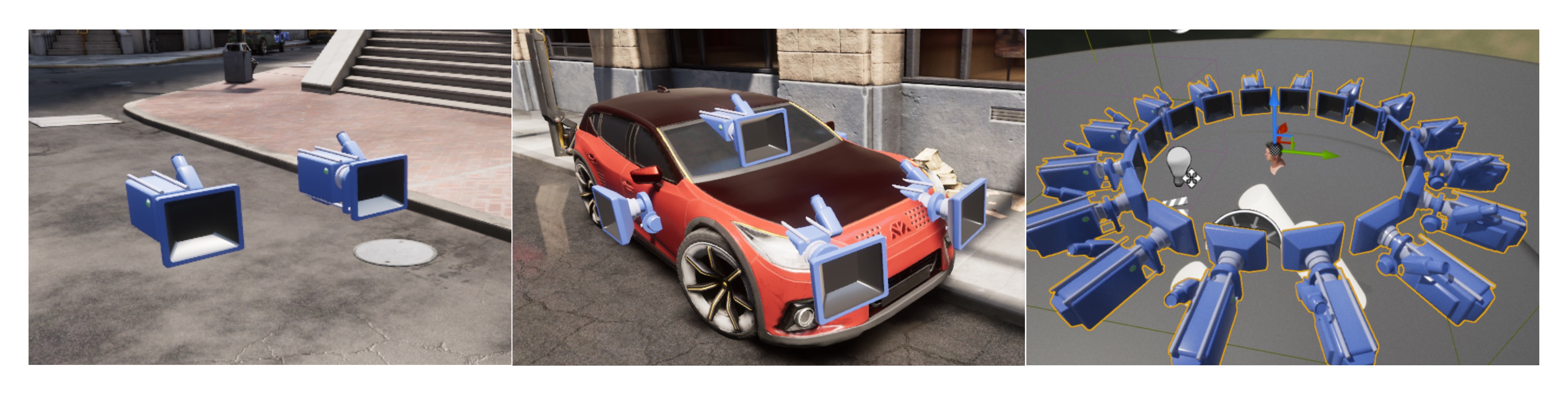}
    \caption{\label{fig:multi_cam} UnrealStereo is a synchronized multiple camera system. From left to right are a two-camera system used in our stereo experiment, cameras mounted on a virtual car and a 16 camera system surrounding a virtual human head.}
\end{figure*}

Game and movie industries are able to create realistic computer graphics images, but it is expensive and technically challenging for researchers to do so. Professional tools such as Blender and Maya are difficult to use because 1) they are created for professional designers with many irrelevant features to research, mastering these tools requires weeks to months experience, 2) they are designed for rendering images and require a significant engineering effort to generate correct ground truth for vision tasks, 3) 3D models for these tools are either expensive or of low-quality. 

UnrealStereo solves these problems by providing an easy-to-use tool. The tool is designed for multi-view vision data generation and diagnosis for researchers. It is based on Unreal Engine 4 (UE4), an open-source 3D game engine.

\begin{figure}
	\centering
    \includegraphics[width=\linewidth]{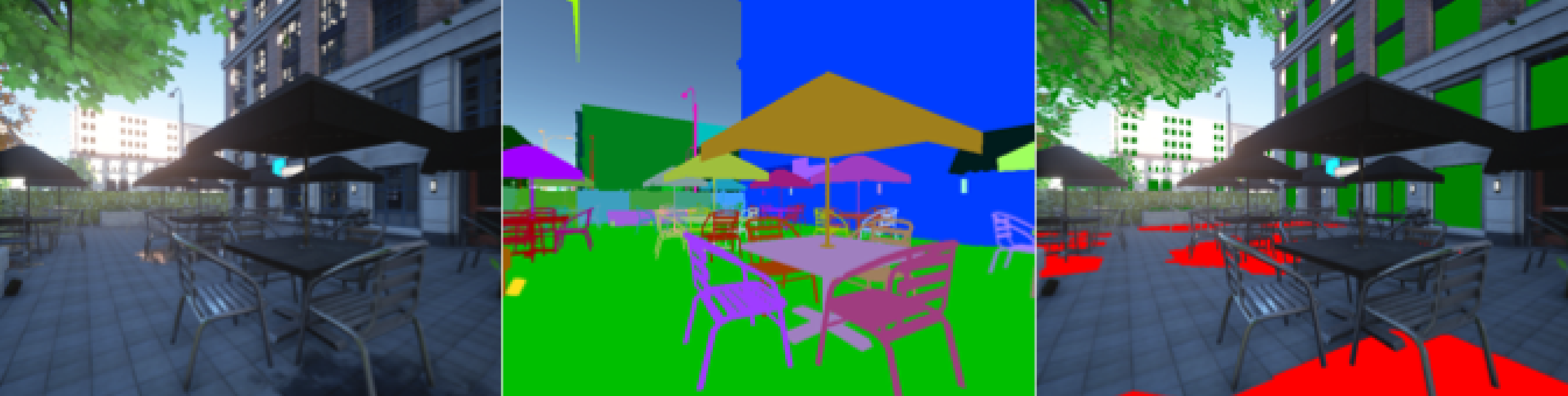}
    \caption{\label{fig:extra_gt} From left to right are rendered images, object instance mask, material information (green shows transparent and red shows specular region).}
\end{figure}

UnrealStereo supports multiple camera. Users can place virtual cameras in a virtual scene according to their specification. An example is shown in Fig.~\ref{fig:extra_gt}. It generates images and ground truth synchronously from multiple cameras, which enables capturing a dynamic scene. Our optimized code makes data generation very fast and only a small overhead is added to the rendering. For a two-camera setup, the speed can reach 30 - 60 FPS depending on complexity of the scene. The speed is important for large scale data generation and interactive diagnosis.

The depth generation from Unreal Engine is improved based on \cite{qiu2016unrealcv}. The depth is stored as floating point instead of 8-bit integer to preserve precision. The depth of transparent objects is missing from the depth buffer of UE4, this issue is fixed to produce accurate depth for transparent objects. Dynamic scenes and visual effects are supported. Many scenes were tested to ensure compatibility. 

For the stereo analysis, we created a two-camera system. The second camera automatically follows the first one and keeps relative position fixed. The distance between two cameras can be adjusted to simulate different baseline. The image and depth are captured from the 3D scenes for both two cameras, along with other extra information shown in Fig.~\ref{fig:extra_gt}. Given a rectified image pair, the goal of stereo matching is to compute the disparity $d$ for each pixel in the reference image. The disparity is defined as the difference in horizontal location of a point in the left image and its corresponding one in the right. Then the conversion between depth $z$ and disparity $d$ is shown in the following relation $ z = \frac{fB}{d} $. where $f$ is the focal length of the camera and $B$ is the baseline that is the distance between the camera centers. The correctness of disparity is verified by warping the reference image according to its disparity map and comparing it with the target image. 

UnrealStereo supports hazardous factor control, such as adjusting material property, which enables the diagnosis experiment in Sec.~\ref{subsection:manual}. The hazardous factor control can be done with Python, through the communication layer provided by UnrealCV~\cite{qiu2016unrealcv}. This makes it possible to generate various cases to stress test an algorithm.

The 3D scenes used in this paper are created by 3D modelers trying to mimic real world configuration. This is important for two reasons: 1) many diverse challenging cases can prevent over-fitting which usually happens in a toy environment. 2) the semantic information provides the opportunity to solve low level vision problems with high level semantic cues~\cite{guney2015displets}. The physics based material system of UE4~\cite{karis2013real} not only makes the rendering realistic, but also enables UnrealStereo to tweak material parameters to create hazardous challenges. 


Unreal Engine uses a rasterization renderer combined with off-line baked shadow map to produce realistic lighting effect. Recently announced V-ray plugin provides another powerful ray tracing renderer for UE4. Our tool can support both renderers. Due to the lack of 3D models for the ray tracing renderer, our synthetic images are mainly produced by the rasterization renderer.

\subsection{Controling Hazardous Factors}
\label{subsection:manual}

\begin{figure*}
\begin{center}
 \subfigure[Specularity]{ \label{fig:subfig:a}  \includegraphics[width=0.4\columnwidth]{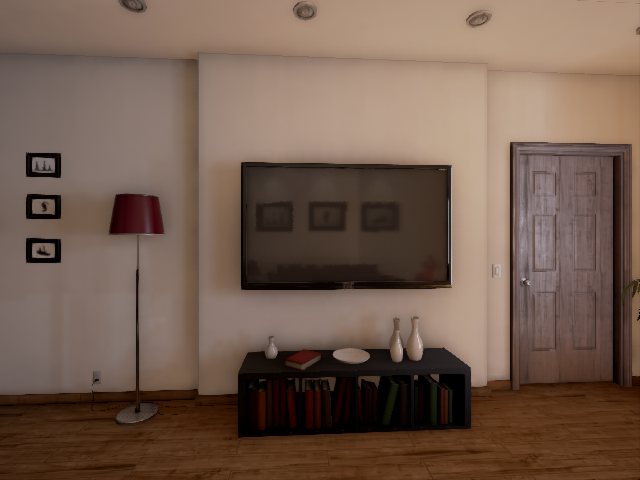}}
 \subfigure[Texturelessness]{\label{fig:subfig:b}  \includegraphics[width=0.4\columnwidth]{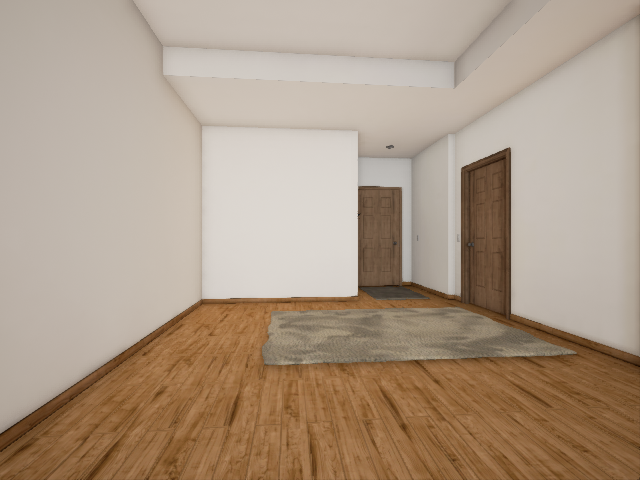}} 
 \subfigure[Transparency]{ \label{fig:subfig:c}  \includegraphics[width=0.4\columnwidth]{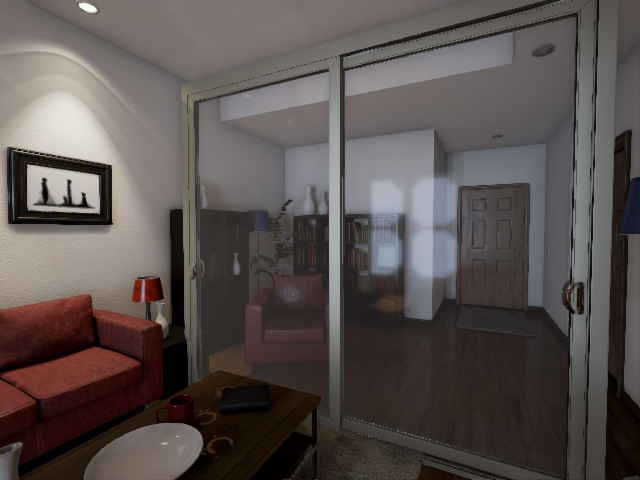}}
 \subfigure[Disparity jumps]{ \label{fig:subfig:d}  \includegraphics[width=0.4\columnwidth]{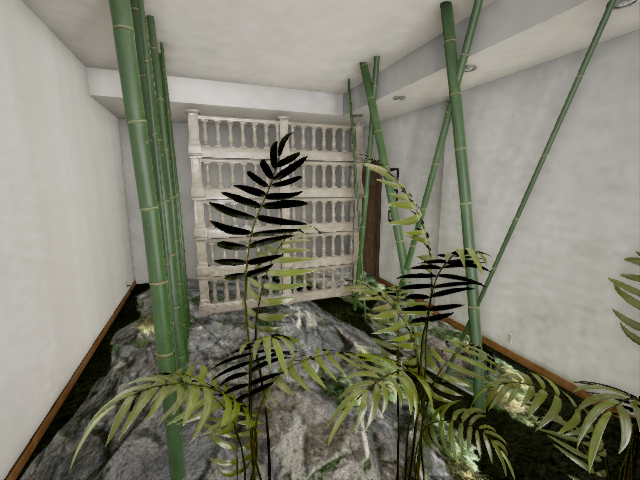}}
 \end{center}
 \caption{\label{fig:cases} From (a) to (d) are cases we designed to test algorithms. They are specularity, lack of texture, transparency and disparity jump. In (a), the screen of a TV is set to be specular. In (b), the wall and the ceiling in the room are made textureless. In (c), the sliding door has a transparent surface. In (d), objects such as bamboos, fences and plants give frequent disparity discontinuities.}
\end{figure*}

The UnrealStereo tool we developed is able to produce hazardous cases in the virtual world with lighting and material controlled, making it tractable to conduct precise evaluation. As a demonstration, we establish four virtual scenes with high reality each of which includes one factor. Stereo image pairs are rendered from various viewpoints together with dense disparity groundtruth. Fig.~\ref{fig:cases} shows the snapshots of the four scenes.

\noindent\textbf{Specularity} 
Shown in Fig.~\ref{fig:subfig:a}, the major specular object in the scene is the TV screen. The specularity is controlled by the roughness of metallic materials.

\noindent\textbf{Texturelessness}
In Fig.~\ref{fig:subfig:b}, the wall and the ceiling in the room are made textureless because they are the most common textureless objects in real world. To achieve texturelessness while keep reality, we do not directly remove the material of the walls but adjust the scale property of the parameterized texture. Various viewpoints are used from which the walls form slanted planes, raising challenges to some less intricate regularizers or smoothness term.

\noindent\textbf{Transparency}
In Fig.~\ref{fig:subfig:c}, we placed a transparent sliding door in a room. By adjusting the opacity property of the glass on the door, we are able to create different levels of transparency.

\noindent\textbf{Disparity Jumps}
In the disparity jumps case(Fig.~\ref{fig:subfig:d}), thin objects such as bamboos, fences and plants of various sizes and poses are placed in the scene, which easily form frequent disparity discontinuities distributed within a small region.

One of the advantage of our tool is the ability to vary the extent of hazard while keeping the rest of the scene intact. We isolate the hazardous factors and focus on one at a time. There are certainly other hazardous factors which can be controlled in our framework. For example, the area of textureless regions is crucial to stereo methods because as the textureless region gets larger, it becomes more difficult for the smoothness term to use context information such as the disparity of neighboring well-textured objects. 



Because synthetic and real data are in different domain, after receiving the evaluation results on virtual scenes, it is important to verify them on real-world dataset. To this end, we manually annotated corresponding hazardous regions on Middlebury 2014~\cite{scharstein2014high} and KITTI 2015~\cite{menze2015object}. Details and results for evaluation on these cases are presented in Section~\ref{experiment:section1}.

\subsection{Automatic Hazardous Region Discovery}

\label{subsection:masks}

Manually designed hazardous cases are important for understanding an algorithm. Furthermore, our tool enables us to tweak many realistic virtual scenes to perform large-scale evaluation. The popularity of virtual reality provides a lot of high quality virtual environments, which can be purchased with a fair price (less than \$50) or even free. 

\begin{figure}
\begin{center}
 \centering
 \includegraphics[width=0.8\columnwidth]{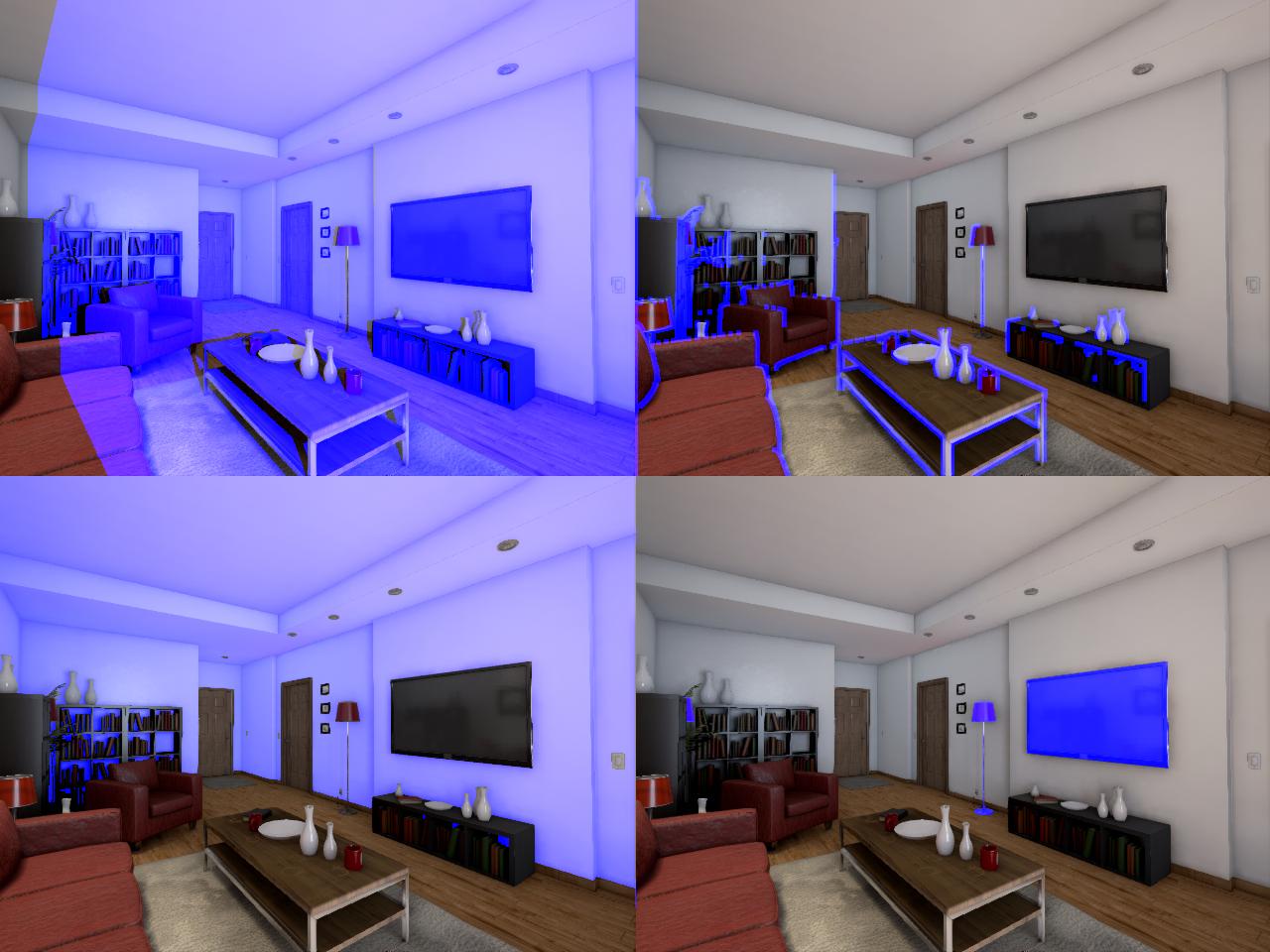}
 \end{center}
 \caption{\label{fig:mask} Binary masks that we compute from object mask and material property. From top left in clockwise are: mask for non-occluded region, object boundary region, specular region and textureless region. Best seen in color.}

\end{figure}

Our rendering process produces extra information beyond depth information including object instance mask and material information. Using these extra information, we can locate these hazardous regions mentioned in Section~\ref{subsection:manual}. Fig.~\ref{fig:mask} shows an example of these masks. For each object, we annotate its material information only once, before rendering process, then no more human effort is required to obtain corresponding masks. 
Textureless regions can also be computed from image using image gradient and disparity jumps regions can be computed given accurate disparity ground truth~\cite{szeliski1999experimental,scharstein2002taxonomy}. Compared with them, our method is a generic way that covers more hazardous factors. 


\begin{figure*}[h]
\centering
\includegraphics[width=0.9\textwidth]{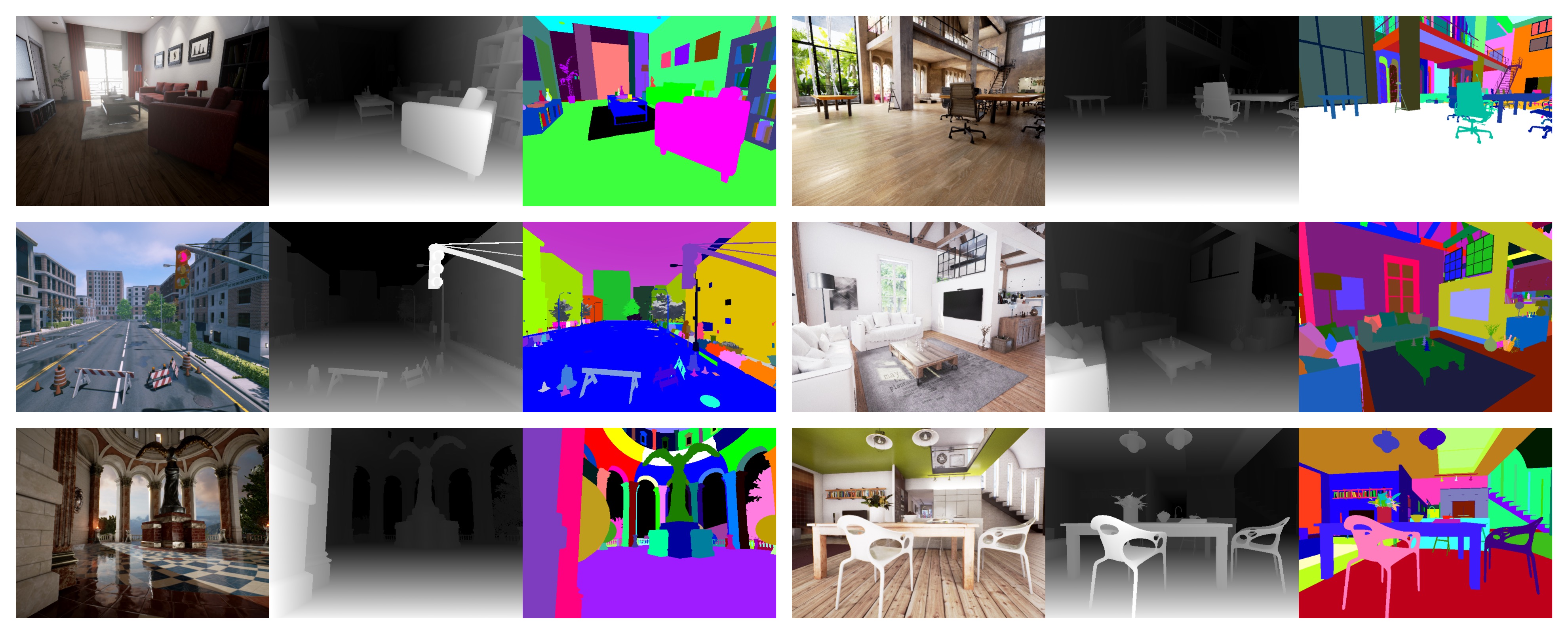}
	\caption{\label{fig:games} The six virtual scenes we use in our experiments, from left to right are image, depth and object mask. These virtual scenes are purchased from Unreal Engine marketplace.}
\end{figure*}

We establish a large dataset using six publicly available game scenes. They are a small indoor room, a large temple scene, three houses and one block of street. There are different layouts in these houses such as living room, kitchen and bathroom. The largest scene contains more than 1,000 objects while hundreds on average, including reflective objects, such as mirrors, bathtubs and metal statues, transparent objects such as glass, glass-doors and windows. Snapshots of these scenes can be seen in Fig.~\ref{fig:games}. Specifically, for each scene we record a video sequence that covers different viewpoints in the environment, which results in 10,825 image pairs in total.

A unique feature of our dataset is the hazardous factors of virtual worlds can be controlled and more challenging images can be produced. 
Instead of just providing an image dataset with fixed number of images, we provide a synthetic image generation tool. This tool can be used to design new hazardous cases, generate more images. More game scenes from the marketplace can be used in experiment.


\section{Experiment}

We choose five types of state-of-the-art stereo algorithms to evaluate on the challenging testing data we rendered. They are representatives of local methods ELAS~\cite{geiger2010efficient} and local method with spatial cost aggregation CoR~\cite{chakrabarti2015low}, global methods on pixel-level MC-CNN~\cite{zbontar2015computing} and superpixel-level SPS-St~\cite{yamaguchi2014efficient} as well as end-to-end CNN based method DispNetC~\cite{mayer2016large}. Implementation from the authors of these methods are adopted. For model weights of the MC-CNN, we use the model used in their submission to KITTI. For DispNetC, the original model trained on the synthetic dataset FlyingThings3D~\cite{mayer2016large} is used. 
Two error metrics, i.e. bad-pixel percentage (BadPix) and end-point error (EPE), are used in evaluation.

\subsection{Evaluation on Controlled Hazardous Levels \label{experiment:section1}}

\begin{figure}[h]
	\includegraphics[width=0.235\textwidth]{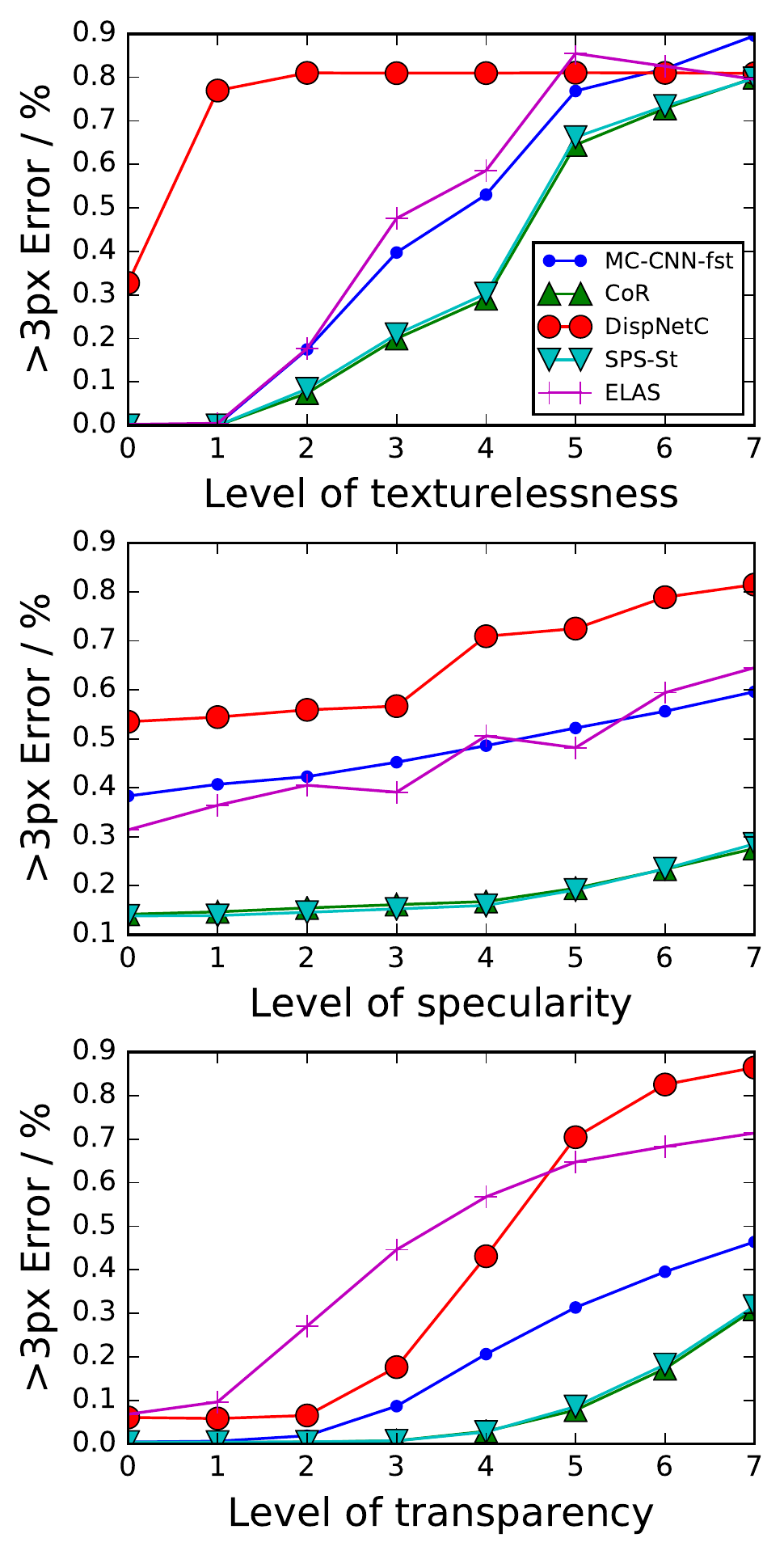}
	\includegraphics[width=0.235\textwidth]{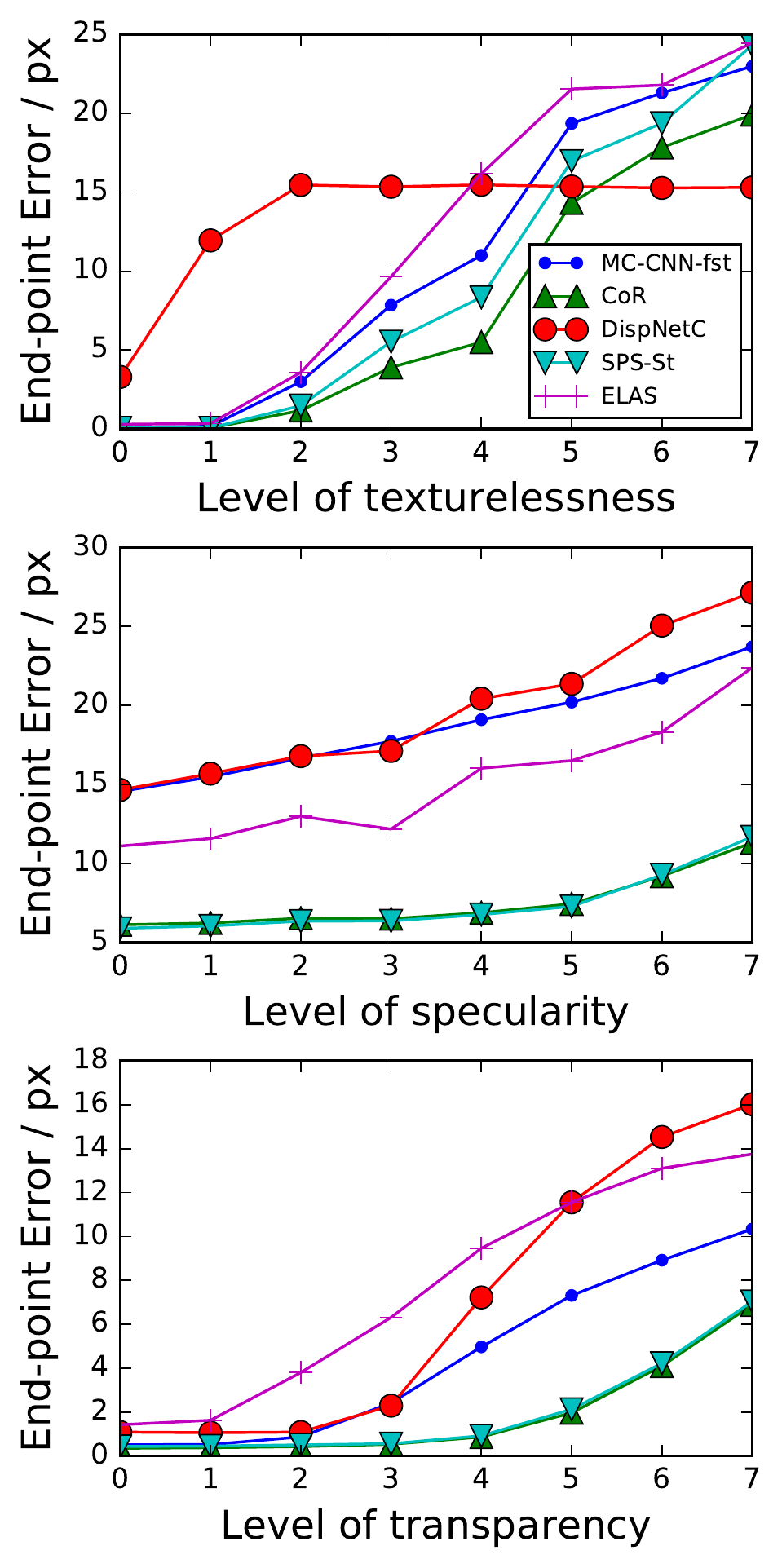}
	\caption{\label{fig:case_performance} The influence of texturelessness, specularity and transparency at different levels in terms of bad-pixel percentage and end-point error. The level of each hazardous factor is controlled by parameters for corresponding materials. Each data point represents an average over 10 viewpoints. }
\end{figure}

We use 10 viewpoints for each of the hazardous cases we designed, i.e. specular, semi-transparent, textureless, and disparity jumps, covering both fronto-parallel and slanted surfaces. At each viewpoint of hazardous scenes except disparity jumps case, we start from the easiest parameter settings that are roughest, opaque or well-textured and adjust the corresponding parameter step by step to increase the extent of hazard, creating different levels of corresponding hazard per viewpoint. We exclude occluded regions and only evaluate hazardous regions identified by method described in Section~\ref{subsection:masks}. Results are shown in Fig.~\ref{fig:case_performance} and Table~\ref{tab:case1}. As a reference, overall performance on Middlebury and KITTI is shown in Table.~\ref{tab:case1}

\begin{table*}[]
\centering\small
\begin{tabular}{m{70pt}||>{\centering\arraybackslash}m{15pt}>{\centering\arraybackslash}m{15pt}>{\centering\arraybackslash}m{15pt}>{\centering\arraybackslash}m{16pt}||>{\centering\arraybackslash}m{15pt}>{\centering\arraybackslash}m{15pt}>{\centering\arraybackslash}m{15pt}>{\centering\arraybackslash}m{16pt}||>{\centering\arraybackslash}m{15pt}>{\centering\arraybackslash}m{16pt}||>{\centering\arraybackslash}c||>{\centering\arraybackslash}m{15pt}>{\centering\arraybackslash}m{16pt}}
\hline
\multirow{2}{*}{} & \multicolumn{4}{c||}{Specular} & \multicolumn{4}{c||}{Textureless} & \multicolumn{2}{c||}{Transparent} & \multicolumn{1}{c||}{Jumps} & \multicolumn{2}{c}{Overall} \\
& \multicolumn{1}{c}{High} & \multicolumn{1}{c}{Med} & \multicolumn{1}{c}{MB} & \multicolumn{1}{c||}{KITTI} & \multicolumn{1}{c}{High} & \multicolumn{1}{c}{Med} & \multicolumn{1}{c}{MB} & \multicolumn{1}{c||}{KITTI} & \multicolumn{1}{c}{Ours} & \multicolumn{1}{c||}{KITTI} & \multicolumn{1}{c||}{Ours} & \multicolumn{1}{c}{MB} & \multicolumn{1}{c}{KITTI} \\ \hline

\multicolumn{14}{c}{$>$3px Error (\%)} \\ \hline

ELAS~\cite{geiger2010efficient} & 64.5 & 39.1 & 57.9 & 35.9 & \textbf{79.5} & 47.6 & 67.4 & 37.6 & 71.4 & 43.3 & 46.4 & 19.5 & 9.3 \\
SPS-St~\cite{yamaguchi2014efficient}   & 28.5 & \textbf{15.3} & 36.5 & 23.0 & 79.8 & 20.9 & 58.3 & 15.0 & 31.6 & 44.6 & 42.3 & 15.8 & 5.1 \\
CoR~\cite{chakrabarti2015low} & \textbf{27.6} & 16.1 & \textbf{30.5} & 23.4 & 79.9 & \textbf{20.0} & 59.0 & \textbf{14.3} & \textbf{30.9} & 44.2 & \textbf{42.1} & 15.4 & 4.9 \\
MC-CNN-fst~\cite{zbontar2015computing} & 59.6 & 45.2 & 37.2 & \textbf{21.2} & 89.5 & 39.7 & 49.1 & 17.4 & 46.4 & \textbf{42.1} & 42.3 & \textbf{14.6} & \textbf{4.5} \\
DispNetC~\cite{mayer2016large} & 81.5 & 56.7 & 82.4 & 36.8 & 80.9 & 81.0 & \textbf{31.8} & 32.0 & 86.5 & 58.2 & 63.4 & 23.5 & 10.7  \\ \hline

\multicolumn{14}{c}{End-point Error (px)} \\ \hline

ELAS~\cite{geiger2010efficient}        & 22.39 & 12.18 & 7.54 & 4.07 & 24.44 & 9.64 & 7.95 & 6.43 & 13.76 & 5.29 & 11.95 & 4.01 & 1.55 \\
SPS-St~\cite{yamaguchi2014efficient}   & 11.69 & \textbf{6.37} & 5.47 & 2.71 & 24.29 & 5.53 & 8.12 & 1.95 & 7.04 & \textbf{5.03} & 11.21 & 3.57 & 1.23 \\
CoR~\cite{chakrabarti2015low} & \textbf{11.32} & 6.50 & \textbf{4.31} & \textbf{2.61} & 19.89 & \textbf{3.90} & 8.05 & \textbf{1.82} & \textbf{6.88} & 5.24 & 11.15 & 3.35 & 1.11 \\
MC-CNN-fst~\cite{zbontar2015computing} & 23.72 & 17.73 & 5.19 & 2.62 & 22.96 & 7.84 & 8.00 & 3.16 & 10.34 & 5.04 & 11.23 & \textbf{3.18} & \textbf{1.10} \\
DispNetC~\cite{mayer2016large} & 27.15 & 17.12 & 8.96 & 3.42 & \textbf{15.30} & 15.33 & \textbf{2.66} & 3.80 & 16.03 & 6.79 & \textbf{10.01} & 3.25 & 1.59     \\ \hline
\end{tabular}
\vspace{5pt}
\caption{ Performance on hazardous regions on images generated by UnrealStereo and corresponding regions on Middlebury (MB) and KITTI training set in bad-pixel percentage (BadPix) and end-point error (EPE). For our data, Hazardous levels of medium (Med) and high (High) are presented for textureless and specular factors. Only masked hazardous regions are evaluated. \label{tab:case1}}
\end{table*}

\begin{table}[h]
\small
\centering
\begin{tabular}{lcccc}
\hline
& Spec. & Txtl. & Tran. & Jumps\\
\hline
KITTI & 0.55 & 0.16 & 0.75 & -  \\
MB & 0.76 & 0.87 & - & -  \\
\hline

\end{tabular}
\vspace{5pt}
\caption{\label{tab:corr} Correlation between performance on our dataset and real-world datasets on hazardous regions in EPE.}
\end{table}

The ability to control hazardous factors enables us to analyze a stereo algorithm from different perspectives. We can study not only the overall performance, but also the robustness to different hazardous cases. Here are some interesting observations from the experiment results.

First, methods which perform better in general are not always doing well on hazardous regions. For example, the state-of-the-art method MC-CNN achieves the best overall scores on both real-world datasets and our synthetic dataset (see Table~\ref{tab:attr}), but it is not the best for many hazardous cases. We compute the correlation coefficients of the performance of these methods for hazardous factors at high level and their overall performance in EPE. For specular, textureless, transparent and disparity jumps factors, they are $0.25$, $0.41$, $0.43$, $0.63$ respectively. Therefore, the overall scores cannot reflect the characteristics of an algorithm on hazardous regions.

Second, different regularization methods have big impact on the robustness. The cost aggregation on suitable regions or regularization on superpixels can to some extent reduce the vulnerability to matching ambiguities. As shown in Fig.~\ref{fig:case_performance}, CoR and SPS-St exhibit high robustness as they outperform other methods for specularity and transparency factors at all levels under both metrics. Intuitively, large support regions also helps regularize the result on textureless regions, which is confirmed by the leading performance of CoR and SPS-St for texturelessness. 


Third, the ability to precisely control the hazardous factors enable us to discover more characteristics of the algorithms than using standard benchmarks. As shown in the curves for textureless in Fig.~\ref{fig:case_performance}, DispNetC exhibits an early insensitivity to further texture weakening, which may result from a different way to incorporate context, i.e. through large receptive field. Without controlling hazardous factors, it is hard to discover these kinds of information. 


From the experiments for disparity jumps, we find that the global methods evaluated here still suffer a lot on these areas even though they have taken depth discontinuity into consideration. The evaluated methods perform bad on disparity discontinuity regions as shown in Table~\ref{tab:case1}. For BadPix metric, CoR is slightly better than others while DispNetC achieves the best result in EPE. The reason that DispNetC outperforms others in EPE could be that it does not explicitly impose smoothness constraints, which helps to avoid erroneous over-smooth.

\subsection{Comparison with Middlebury and KITTI\label{experiment:compare}}

\begin{figure}[h]
	\centering
  \includegraphics[width=0.8\columnwidth]{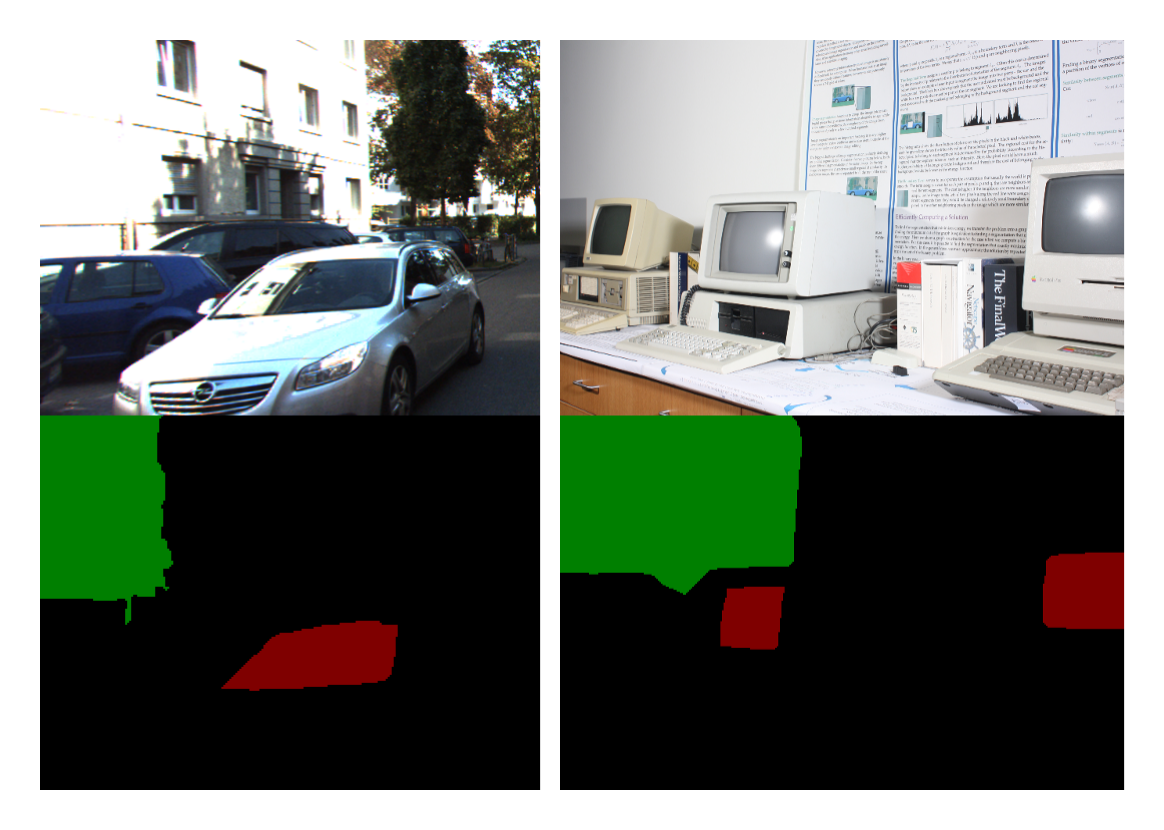}
	\caption{\label{fig:KITTI_MB_anno} Hazardous regions annotation on KITTI (left) and Middlebury (Right) used to validate the results on synthetic data. Specular and textureless regions are encoded by red and green color.} 
\end{figure}

To verify our result, we annotate specular and textureless regions on Middlebury 2014 and KITTI 2015 training set and transparent regions on the latter (Note that the objects in Middlebury are rarely transparent). On Middlebury the annotation and evaluation are performed at quarter size of the original images. Disparity jumps is not included here because the missing ground truth for many pixels on both datasets makes disparity discontinuity computation inaccurate. To annotate hazardous regions of these datasets, annotators are asked to mask corresponding regions with Photoshop selection tool and examples are shown in Fig~\ref{fig:KITTI_MB_anno}. 

\begin{table*}[h]
\centering
\begin{tabular}{@{}l|>{\centering\arraybackslash}c>{\centering\arraybackslash}c|>{\centering\arraybackslash}c>{\centering\arraybackslash}c|>{\centering\arraybackslash}c>{\centering\arraybackslash}c|>{\centering\arraybackslash}c>{\centering\arraybackslash}c|>{\centering\arraybackslash}c>{\centering\arraybackslash}c|>{\centering\arraybackslash}c>{\centering\arraybackslash}c@{}}
\hline
 & \multicolumn{2}{c|}{Full} & \multicolumn{2}{c|}{Non-Occluded} & \multicolumn{2}{c|}{Specular} & \multicolumn{2}{c|}{Textureless} & \multicolumn{2}{c|}{Transparent} & \multicolumn{2}{c}{Disparity jumps}\\
 & \multicolumn{1}{c}{EPE} & \multicolumn{1}{c|}{\textgreater3px} & \multicolumn{1}{c}{EPE} & \multicolumn{1}{c|}{\textgreater3px} & \multicolumn{1}{c}{EPE} & \multicolumn{1}{c|}{\textgreater3px} & \multicolumn{1}{c}{EPE} & \multicolumn{1}{c|}{\textgreater3px} & \multicolumn{1}{c}{EPE} & \multicolumn{1}{c|}{\textgreater3px} & \multicolumn{1}{c}{EPE} & \multicolumn{1}{c}{\textgreater3px} \\ \hline
ELAS~\cite{geiger2010efficient} & 11.80 & 31.6 & 8.81 & 25.1 & 8.18 & 22.6 & 14.04 & 56.0 & 11.37 & 42.2 & 10.21 & 43.0 \\
SPS-St~\cite{yamaguchi2014efficient} & 7.93 & 23.4 & 5.16 & 16.3 & \textbf{6.46} & \textbf{16.3} & 10.74 & 44.2 & \textbf{9.94} & \textbf{36.9} & 6.72 & 30.7 \\
CoR~\cite{chakrabarti2015low} & 7.74 & 24.2 & 4.97 & 16.9 & 7.07 & 24.4 & 8.34 & 45.8 & 10.07 & 37.1 & 6.70 & 32.0 \\
MC-CNN-fst~\cite{zbontar2015computing} & \textbf{7.64} & \textbf{22.2} & \textbf{4.62} & \textbf{14.5} & 6.94 & 17.1 & 7.62 & 41.2 & 10.52 & 37.0 & \textbf{6.56} & \textbf{30.5} \\
DispNetC~\cite{mayer2016large} & 7.98 & 34.7 & 5.96 & 28.4 & 7.84 & 29.3 & \textbf{6.02} & \textbf{37.3} & 12.76 & 52.7 & 6.94 & 44.8 \\
 \hline

\end{tabular}
\vspace{5pt}
\caption{\label{tab:attr} Performance of state-of-the-art stereo algorithms on test set of rendered dataset. Errors in full image, non-occluded, specular, textureless, transparent and disparity jumps regions are included. Both end-point error (EPE) and bad-pixel percentage ($>3$px) are evaluated by applying the masks proposed in Section~\ref{subsection:masks}. }
\end{table*}

Performance on annotated hazardous regions is consistent with our synthetic dataset. As shown in Table~\ref{tab:corr}, there is a strong correlation between performance on our dataset and real-world datasets. For textureless regions on KITTI, the correlation coefficient is $0.16$ at high level and $0.54$ for medium level, which indicates that KITTI shares similar statistics for textureless regions with our dataset at medium level. 

As shown in Table~\ref{tab:case1}, MC-CNN does not outperform others on hazardous regions on Middlebury and KITTI, which verifies the first conclusion in Section~\ref{experiment:section1} that methods which perform better in general are not always doing well on hazardous regions. The second conclusion also holds true here. Since global methods, e.g. SPS-St and MC-CNN, and local methods with large support regions, e.g. CoR, obtain lower errors on specular and transparent regions than other methods, they are more robust to these hazardous factors.

We also find that Middlebury and KITTI have different statistics. For example, on textureless regions, DispNetC performs the best on Middlebury while on KITTI it does not. The analysis of DispNetC in Sec.\ref{experiment:section1} shows it has different performance for different levels of texturelessness. Since Middlebury and KITTI are both real-world dataset and the level of hazardous factors is unknown and not controllable, the performance for DispNetC can be different. According to Fig.~\ref{fig:case_performance}, it is possible that the annotated textureless regions on Middlebury are at the higher level while those on KITTI is more towards the lower level.

\subsection{Evaluation on Automatically Generated Hazardous Regions \label{experiment:section2}}


We evaluate these algorithms on a testing set including 484 stereo image pairs which are randomly sampled from the 10k images from the six virtual scenes. Hazardous regions are generated automatically. The average performance on full, non-occluded and four hazardous regions are shown in Table.~\ref{tab:attr}. 


The top performance of SPS-St and CoR on specular and transparent regions verifies the analysis in Section~\ref{experiment:section1} that non-local regularization using large support regions would reduce the influence of matching ambiguity. That DispNetC outperforms others on textureless region could result from the level of texturelessness, since Fig.~\ref{fig:case_performance} shows that DispNetC is robust on extremely textureless scene. 

It is also worthwhile to compare the results with overall performance on Middlebury and KITTI in Table.~\ref{tab:case1}. The correlation coefficients for the performance in EPE between our dataset and Middlebury and KITTI are 0.91 and 0.61 respectively. The overall errors are higher on our data. There are two possible causes. One is that the percentage of hazardous regions on our dataset is larger than KITTI. The other is that KITTI only provides semi-dense ground truth, which excludes many hazardous regions, i.e. the windows of cars.

\section{Conclusion}

In this paper, we presented a data generation tool UnrealStereo to generate synthetic images to create a stereo benchmark. We used this tool to analyze the effect of four hazardous factors on state-of-the-art algorithms. Each factor was varied at different degrees and even to an extreme level to study its impact. We also tested several state-of-the-art algorithms on six realistic virtual scenes. The hazardous regions of each image were automatically computed from the ground truth, e.g., the object mask and the material properties. We found that the state-of-the-art method MC-CNN~\cite{zbontar2015computing} outperforms others in general, but lacks robustness in hazardous cases. DCNN based method~\cite{mayer2016large} exhibits interesting properties due to its awareness of larger context. We also validated our findings by comparing to results on the real-world datasets where we manually annotated the hazardous regions. The synthetic data generation tools enables us to explore many degrees of hazardous factors in a controlled setting, so that the time-consuming manual annotation of real images can be reduced. Manual annotation will only be needed in a limited (sparse) number of cases in order to validate the results from synthetic images. 

Our data generation tool can be used to produce more challenging images and is compatible with publicly available high-quality 3D game models. This makes our tool capable for many applications other than stereo. In our future work, we will extend our platform to include more hazardous factors such as the ratio of occlusion and analyze more computer vision problems. It is also interesting to explore the rich ground truth we generate, such as object mask and material properties. This semantic information will enable the development of computer vision algorithms that utilizes high-level knowledge, for example like stereo algorithms that use 3D car models~\cite{guney2015displets}.

\noindent\footnotesize\textbf{Acknowledgement:} This work was supported by the Intelligence Advanced Research Projects Activity (IARPA) via Department of Interior/ Interior Business Center (DOI/IBC) contract number D17PC00345. We also want to thank the reviewers for providing useful comments.

{
\small
\bibliographystyle{ieee}
\bibliography{stereo}
}

\end{document}